\documentclass[]{article}

\pdfoutput=1

\usepackage{multirow}
\usepackage{graphicx}
\usepackage{amsmath}
\usepackage{authblk}

\title{Predicting Job-Hopping Motive of Candidates Using Answers to Open-ended Interview Questions}

\author{Madhura Jayaratne}
\author{Buddhi Jayatilleke}
\affil{\scriptsize{PredictiveHire Pty. Ltd., 15, Newton Street, Cremorne, VIC 3121, Australia}}
\date{}

\begin{document}
	
\maketitle

\begin{abstract}
A significant proportion of voluntary employee turnover includes people who frequently move from job to job, known as job-hopping. Our work shows that language used in responding to interview questions on past behaviour and situational judgement is predictive of job-hopping motive as measured by the Job-Hopping Motives (JHM) Scale. The study is based on responses from over 45,000 job applicants who completed an online chat interview and self-rated themselves on JHM Scale. Five different methods of text representation were evaluated, namely four open-vocabulary approaches (TF-IDF, LDA, Glove word embeddings and Doc2Vec document embeddings) and one closed-vocabulary approach (LIWC). The Glove embeddings provided the best results with a correlation of r = 0.35 between sequences of words used and the JHM Scale. Further analysis also showed a correlation of r = 0.25 between language-based job-hopping motive and the personality trait \textit{Openness to experience} and a correlation of r = -0.09 with the trait\textit{ Agreeableness}.
\end{abstract}


\section{Introduction}
\label{sec:introduction}

Voluntary turnover, which represents the vast majority of all employee turnover, decreases organizational productivity and dampen employee morale \cite{lam_new_2002} while inflicting direct financial costs related to rehiring such as sourcing, recruiting and onboarding. However making frequent voluntary job changes, known as job-hopping, has become a trend in the recent past \cite{Chatzky_2018}. The motivations for job-hopping, have been identified to be two-fold; advancement and escape \cite{lake_validation_2018}. The advancement motive represents the growth and career perspective, while the escape motive represents a withdrawal or dislike of the work environment, especially among those who are described as impulsive and unpredictable. The latter is identified as a psychological property and commonly known as the ``hobo syndrome'' \cite{Ghiselli_1974}. Further studies have shown the relationship between personality and voluntary turnover \cite{barton_personality_1972, jenkins_self-monitoring_1993, salgado_big_2002, timmerman_predicting_2006, zimmerman_understanding_2008}.

The ability to assess a candidate's motivation for job-hopping prior to selection can help both candidates and employers make better decisions and avoid future surprises and costs due to voluntary exits. The most frequently used approach for discovering patterns of job-hopping is to explore the employment history listed in an applicant's resume. Sifting through resumes can be both time-consuming and unreliable, especially in situations of high volume recruitment. Resumes are also known to produce biased outcomes \cite{NBERw9873, Kang_2016}. Moreover, it is an ineffective method with novice job seekers, such as new graduates with insignificant job histories. 

In this study, we examine whether answers given by candidates to interview questions related to past behaviour and situational judgement demostrate a correlation to their job-hopping motives as measured by the Job-Hopping Motives (JHM) Scale \cite{lake_validation_2018}, a validated self-report measure of job-hopping motives. The basis for selecting interview answers as a possible predictor is two-fold. Firstly, one's language use has been shown to be highly predictive of their personality. Personality traits have been successfully derived from informal (microblogs \cite{golbeck_predicting_2011, sumner_predicting_2012, xue_personality_2017}, social media posts \cite{tadesse_personality_2018, wang_smotetomek-based_2019}), semiformal (blogs \cite{iacobelli_large_2011}, interview questions \cite{Madhura_2020}) and formal (essays \cite{neuman_vectorial_2014}) contexts. Authors own prior work has shown that interview answers are a strong predictor of personality traits \cite{Madhura_2020}. Secondly, structured interviews where the same questions are asked from every candidate in a controlled conversation flow and evaluated using a well-defined rubric have shown to reduce bias \cite{levashina_2014} and also increase the ability to predict future job performance \cite{mcdaniel_whtzel_schmidt_1994}. Computational inference of job-hopping motive from interview responses further increases the utility of the structured interview and its applicability in high volume recruitment.

In this work, we make the following contributions to the crossroads of computational linguistics and organizational psychology.
\begin{enumerate}
	\item We demonstrate that responses to typical interview questions related to past behaviour and situational judgement can be used to reliably infer one's job-hopping motive as measured by the JHM Scale.
	\item We evaluate multiple methods of text representations and establish that the Glove based word-embedding method achieves the highest correlation of r=0.35 between text and JHM Scale when used with a Random Forest regressor.
	\item We validate the positive correlation between job-hopping motive and \textit{Openness to experience} (one of the personality traits in the HEXACO personality model), both derived from text (r=0.25). This is in line with previous findings using standard personality tests.
\end{enumerate}

The rest of the paper is organised as follows. Section \ref{sec:background} presents a detailed background into the research on employee turnover, the role of personality on turnover and the link between language and personality. In section \ref{sec:methodology}, we describe the methodology, including the data used and the five different text representation methods we evaluated, namely TF-IDF, LDA, Glove word embeddings, Doc2Vec document embeddings and LIWC. Results, in terms of the accuracies achieved by each text representation method, are presented in section \ref{sec:results} along with discussion and further analysis of salient correlations, demographics, and terms used. Section \ref{sec:conclusion} concludes the paper with a summary and future research directions.

\section{Background}
\label{sec:background}

A study conducted by the Australian HR Institute in 2018 across all major industry sectors in Australia \cite{begley_turnover_2018} found that on average companies face an annual turnover rate of 18\% and within the age group of 18 to 35 it jumps to 37\%. That is more than 1 in 3 people in the youngest age group leaving an organization within a year. A majority (63\%) of respondents in the study, mostly HR staff, claimed that their organisation does not measure the financial cost of employee turnover. Employee turnover rate is much higher than the average for some industries such as hospitality \cite{cho_measuring_2006, lam_new_2002}. Cho et al. \cite{cho_measuring_2006} report a staggering 115\% turnover rate among non-managerial employees of hospitality firms while for managerial employees it is 35\%.

Majority of employee turnover consists of voluntary turnover, that is, employee-initiated separations compared to involuntary turnover initiated by the employer such as layoffs and terminations due to poor performance. Significant costs have to be borne by an organization when an employee voluntarily leaves. These include replacement costs such as costs associated with advertising, screening and selecting a new candidate, employee training costs, and operational efficiency losses until a new employee reaches a sufficient level of productivity. These costs are exacerbated by the dampening of remaining employee morale leading to lower quality work and lower productivity \cite{lam_new_2002}. A study conducted by the Work Institute in the US \cite{mahan_retention_2018} found that voluntary exit of an employee costs a company 33\% of the employee's base salary, which the authors claim is a conservative estimate. The report also states that with a median base salary of \$45,000, it is costing the US economy close to \$600bn a year due to voluntary turnover. Similarly, in a large-scale meta-analysis (N $>$ 300,000), Park and Shaw \cite{park_turnover_2013} observed a significant and negative ($\rho = -0.15$) correlation between voluntary turnover rates on organizational performance. 

Voluntary turnover has been associated with a number of negative job attributes such as low level of job satisfaction, lack of promotion opportunities, lack of work-life balance, lack of fairness of the firm's procedures etc. \cite{harman_psychology_2007}, which are reasons originating from a misalignment between employee and employer expectations. Measures can be put in place by the employer to discover and address these issues where possible and methods such as employee engagement surveys and periodic review discussions serve this purpose. 

The focus of our work is a type of voluntary turnover identified as ``job hopping'', the frequent move from job to job \cite{lake_validation_2018} by some individuals than others. Studies have shown that there is a relationship between one's personality and job-hopping motives \cite{barton_personality_1972, jenkins_self-monitoring_1993, ariyabuddhiphongs_big_2015, salgado_big_2002, timmerman_predicting_2006, zimmerman_understanding_2008, hong_relationship_2008, sarwar_study_2013}. The hypothesis behind these studies is that one's personality plays a key role in their intention to hop jobs and acts as a latent variable that mediates their desire to voluntarily leave an organization. Barton and Cattell \cite{barton_personality_1972} conducted one of the earliest longitudinal studies on the effect of personality on job promotion and job change and found that individuals who were more \textit{practical} and \textit{down to earth} recorded the lowest incidents on job turnover. Ghiselli \cite{Ghiselli_1974} named this tendency the ``hobo syndrome'', an internal impulse-driven action to move from one job to another shown by some employees irrespective of other more rational motives. Lake et al \cite{lake_validation_2018} refer to this as an ``escape motive'' or a sudden withdrawal from the work environment as opposed to an ``advancement motive'' that makes an employee leave for a perceived better opportunity. More recent studies have also shown a correlation between hobo syndrome and the Big 5 personality trait of \textit{Openness to experience} \cite{woo_2011} further validating the personality influence on job-hopping motives. Other studies have shown similar relationships between the Big-5 personality traits and turnover intention. For instance, Sarwar et at. \cite{sarwar_study_2013} found personality traits \textit{Extraversion}, \textit{Neuroticism}, \textit{Conscientiousness} and \textit{Agreeableness} to be negatively correlated with turnover intention while \textit{Openness to experience} was positively related with turnover intention. These results are in line with the Big-5 personality traits' correlations with the intention to quit observed in a meta-analysis conducted by Zimmerman \cite{zimmerman_understanding_2008}. In a study conducted with call centre employees, Timmerman \cite{timmerman_predicting_2006} reported slightly different results with \textit{Neuroticism}, \textit{Conscientiousness} and \textit{Agreeableness} as negatively correlated with turnover while \textit{Extraversion} and \textit{Openness to experience} positively correlated with turnover. With the above studies, it is pertinent to conclude that overall there is a latent relationship between one's personality and his/her turnover intentions. 

Individual job-hopping motive can be measured using self-rating questionnaires similar to standard personality tests. One such validated self-rating item list is the Job-Hopping Motives Scale developed by Lake, Highhouse and Shrift \cite{lake_validation_2018}. It includes eight self-rating items with four items each validated with factor analysis to assess escape and advancement motives. Their study also confirms the positive correlation of escape motive with impulsivity (r=0.19) and a negative correlation with persistence (r=-0.16). They found the number of jobs the participants had voluntarily quit during their lifetime to be significantly related to both escape and advancement motives (r=0.08, p\textless 0.05 and r=0.10, p\textless 0.01). They also found the ratio of voluntarily quit jobs relative to the total number of jobs held over one’s life to be significantly related to both motives (r=0.11, p\textless 0.01 and r=0.09, p\textless 0.05). Further, assessing whether job-hopping motives could predict work history variables above and beyond established predictors, they found escape motive was significant with $ \beta $=0.14 (p=0.02) in two separate regression models while the advancement motive was not.

However a challenge with administering self-rating based assessments is the need to have multiple statements to gain a measure of a single personality construct (for example four items are required to measure escape motive). It is not unusual to have over 100 items in such tests when you combine other measures such as personality traits. Applicant reactions to such personality tests have shown to be less favourable than interviews \cite{anderson_salgado_2010, hausknecht_day_2004}. Based on a meta-analysis of multiple studies on applicant reaction to selection methods, Anderson et al. \cite{anderson_salgado_2010} found that compared to job interviews and work sample tests, personality tests fall short of making a positive impression with candidates in areas of face validity, opportunity to perform, interpersonal warmth and respectful of privacy. These indicate candidates' preference to express themselves and not be restricted to self-rating themselves on a pre-defined set of multiple-choice questions (typically over 100 items) as found in standard personality tests.

On the other hand, researchers have demonstrated that one's language use is indicative of his/her personality attributes \cite{pennebaker_linguistic_1999, fast_personality_2008, gill_what_2009, hirsh_personality_2009, qiu_you_2012}. This allows structured interviews, which are much more engaging for applicants and permit open expression of applicant thoughts to be used as a source for inferring personality attributes. Authors have demonstrated in \cite{Madhura_2020} how responses to open-ended interview questions can be used to reliably infer one's personality attributes based on the six-factor HEXACO model \cite{ashton_empirical_2007}. 

Combining the relationship between job-hopping motives and personality with that of personality and language use, we hypothesised that one's language use is closely associated with their job-hopping motives. In other words, the the job-hopping motive can be inferred from various characteristics of their language use. Based on this hypothesis, we envisioned building a language-based model that is able to predict a candidate's job-hopping motives as measured by the JHM Scale from answers to regular open-ended interview questions. Such a model would have wide applicability in digitised forms of interviews, be it chat-based, voice or video where the textual content of the candidate answers can be used to infer JHM Scale scores in addition to the personality and communication skills that can be derived from text. This enables the use of digitised interviews, which are much more engaging than standard personality tests and preferred by candidates to be used as a multi-measure assessment scalable to high applicant volumes supported by algorithmic inferences. 

\section{Methodology}
\label{sec:methodology}

In order to test the correlation between language use and job-hopping motive, we built a regression model to infer the JHM Scale rating (discussed above) using textual answers to open-ended interview questions. Given the importance of numerical representation of language in building a machine learning model, we compared the performance of five different text representation methods namely, terms (TF-IDF), topics (LDA), Glove word embeddings, Doc2Vec and LIWC. In this section, we describe the training dataset, the five different text representation methods and the regression model building approach.

\subsection{Data}
We analysed free-text responses from 45,899 candidates who used the PredictiveHire\footnote{https://www.predictivehire.com/} FirstInterview platform, an online chat-based interview tool. Job applicants answer 5-7 open-ended questions and self-rating questions based on a proprietary personality inventory that also included the JHM Scale items designed by Lake et al. \cite{lake_validation_2018}. FirstInterview is typically the very first engagement the applicant has with the hiring organisation, placed at the top of the recruitment funnel and close to 40\% of applicants complete it on a mobile. 

The online interview questionnaire includes open-ended free-text questions on past experience, situational judgement and values. The questions are customisable by role family (e.g. sales, retail, call centre etc.) and specific customer value requirements.

\begin{itemize}
	\item \emph{What motivates you? What are you passionate about?}
	\item \emph{Not everyone agrees all the time. Have you had a peer, teammate or friend disagree with you? What did you do?}
	\item \emph{Give an example of a time you have gone over and above to achieve something. Why was it important for you to achieve this?}
	\item \emph{Sometimes things don't always go to plan. Describe a time when you failed to meet a deadline or personal commitment. What did you do? How did that make you feel?}
	\item \emph{In sales, thinking fast is critical. What qualifies you for this? Provide an example.}
\end{itemize}

The length of textual responses in terms of words had a $\mu = 234.8$ and $\sigma = 212.2$.

The JHM Scale items consist of the following eight statements with a 5-point response scale ranging from \textit{Strongly Agree} to \textit{Strongly Disagree}. 

\begin{itemize}
	\item Because working for one company tends to create boredom, people should move from company to company often.
	\item Even if someone has changed jobs several times, they should take a new job if it involves moving to a better position.
	\item Frequently moving between jobs is perfectly justified when each job change leads to a more impressive job.
	\item When a person discovers they dislike their coworkers, they should move to another job, and keep switching jobs until they finally find a good place to work.
	\item Becoming disinterested in a job is a good reason to move from job to job as often as desired.
	\item It is desirable to regularly move from job to job, looking for the job that best improves one's lifestyle.
	\item Repeatedly changing jobs is an ideal way to get a variety of job experiences.
	\item People should be willing to change jobs as many times as necessary to get the best job possible.
\end{itemize}

\begin{figure}[h]
	\centering
	\includegraphics[width=10cm]{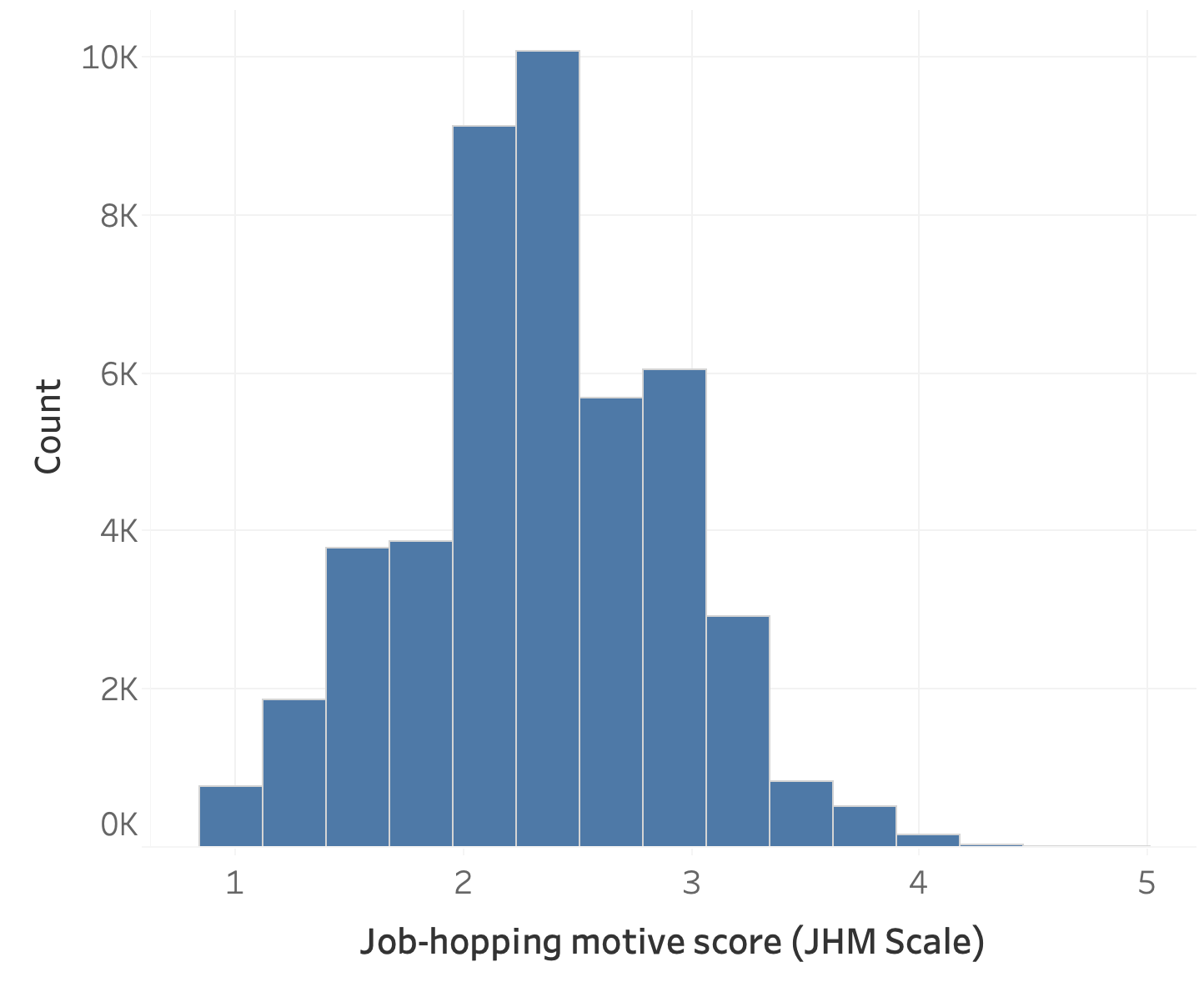}
	\caption{Distribution of job-hopping motive measured by the JHM Scale among all participants}
	\label{fig:distibution}
\end{figure}

Each candidate responded to at least 6 such statements as part of a 40 item personality test. These answers were coded 1 (less likely) to 5 (highly likely)  and a measure on job-hopping motive was formed by averaging over all the questions.  Figure \ref{fig:distibution} shows the distribution of job-hopping motive score ($\mu = 2.343, \sigma = 0.584$) among all participants. This score formed the ground-truth for building the predictive model. The demographics of the candidates in terms of gender and the job role applied are shown in Table \ref{table:table_demographics}.

\begin{table}
	\centering
	\caption{Demographic breakdown of the participants in terms of gender and job family they applied to}
	\label{table:table_demographics}
	\setlength{\tabcolsep}{3pt}
	\begin{tabular}{|p{60pt}|p{85pt}|p{70pt}|}
		\hline
		Attribute & Group & Count\\
		\hline
		\multirow{3}{*}{Gender} 
		& \multicolumn{1}{l}{Female} & \multicolumn{1}{|r|}{7,801} \\
		& \multicolumn{1}{l}{Male} & \multicolumn{1}{|r|}{9,242} \\
		& \multicolumn{1}{l}{Not specified} & \multicolumn{1}{|r|}{28,856} \\
		\hline
		\multirow{6}{*}{Job family} 
		& \multicolumn{1}{l}{Cabin crew} & \multicolumn{1}{|r|}{5,066} \\
		& \multicolumn{1}{l}{Call centre} & \multicolumn{1}{|r|}{1,587} \\
		& \multicolumn{1}{l}{Healthcare} & \multicolumn{1}{|r|}{16,305} \\
		& \multicolumn{1}{l}{Retail} & \multicolumn{1}{|r|}{14,241} \\
		& \multicolumn{1}{l}{Sales} & \multicolumn{1}{|r|}{7,445} \\
		& \multicolumn{1}{l}{Other} & \multicolumn{1}{|r|}{1,255} \\
		\hline
	\end{tabular}
\end{table}

\subsection{Text Representation}

We evaluated four open-vocabulary approaches for representing textual information. Open-vocabulary approaches do not rely on a priori word or category judgments compared to closed vocabulary appraoches, that use predetermined sets of words (dictionaries or lexicons). With the recent advancements in Natural Language Processing (NLP), open-vocabulary approaches have gained popularity and shown better results \cite{gill_what_2009, hirsh_personality_2009, schwartz_personality_2013} over closed-vocabulary approaches such as LIWC \cite{pennebaker_development_2015} used in the past for inferring personality from text. For comparison, we also trained a model using word categories in LIWC, the most commonly used lexicon for text analysis in the psychology domain.\\

\noindent Below we outline the five different text representation methods we used.

\subsubsection{TF-IDF}
Term Frequency-Inverse Document Frequency (TF-IDF) \cite{christopher_introduction_2008} approach uses the relative frequency of occurrence of terms in the text corpus to model the language use. 
That is, the higher usage of a term in a response is scored high while offsetting for the number of responses the term occurs in. More formally, with  $ t $, $ r $, and $ R $ denoting term, response and the set of all responses respectively, $ n_{t,r} $, the number of times term $ t $ appearing in response $ r $ and $ n_{t} $, the number of responses where term  $ t $ appears,
\begin{equation}
tfidf(t, r, R) = tf(t, r) \cdot idf(t, R);~~~~~ t \in r, r \in R
\label{eq:tfidf}
\end{equation}
where
\begin{equation}
tf(t, r) = \dfrac{n_{t,r}}{ \sum_{\acute{t} \in r} n_{\acute{t},r} }
\label{eq:tf}
\end{equation}
\begin{equation}
idf(t, R) = \log(\dfrac{|R|}{n_{t} + 1}) +  1
\label{eq:idf}
\end{equation}

We first tokenized the text responses from interview questions and developed a vectorized representation with the above TF-IDF scheme in n-dimensional space using unigrams, bigrams and trigrams of tokens. We experimented with n-dimensions=500, 1000 and 2000 of the most frequent n-grams (n=1,2,3) being used in the representation and found that n-dimensions=2000 to give the best outcomes.

\subsubsection{LDA}
\label{sec:lda}
Latent Dirichlet Allocation (LDA) \cite{blei_latent_2003} is a topic modelling approach that generates a given number of latent topics from a text corpus. LDA is a generative statistical model which assumes that a document (in our case a candidate response) relates to a number of latent topics while each latent topic is distributed across the vocabulary with different levels of affinities. Hence, a topic is usually described by the terms that have the highest affinities to that topic.

\noindent Using the notation defined in (\ref{eq:tfidf}) and $ \theta $ denoting a LDA topic,
\begin{equation}
p(\theta|r) = \sum_{t \in \theta} p(\theta|t) \times p(t|r)
\label{eq:lda}
\end{equation}

We used the Gensim\footnote{https://radimrehurek.com/gensim/} software package for deriving 100 such topics. An example of a topic derived given by the terms with the highest affinities are \{\textit{food, kitchen, restaurant, cleaning, chef, hospitality, worked, cooking, job}\}. It is important to note that the derivation of coherent topics such as the above is purely based on the statistical distributional properties of the terms in the text corpus.

\subsubsection{Word Embeddings}
The word embedding approaches \cite{mikolov_distributed_2013, jeffrey_pennington_glove_2014} to modelling language derives n-dimensional vectors to represent terms found in a given corpus, a numerical representation that preserves the contextual similarities between words. That is, similar words are placed closer to each other in the vector space. Hence word embeddings can be manipulated and made to perform tasks such as finding the degree of similarity between two words using intuitive arithmetic operations on the word vectors while retaining semantic analogies such as $ woman + king - man = queen $ etc.
Word embeddings based textual representations have been used in solutions that have achieved state-of-the-art results in many NLP tasks \cite{Camacho_2018}.

Word embeddings models are usually trained on large corpora such as Wikipedia or web pages gathered by a web crawler. We used the word embedding model available in the Spacy software package\footnote{https://spacy.io/}, which is trained using the Glove algorithm \cite{jeffrey_pennington_glove_2014} on content from common web crawl. To achieve a vector representation for a given response, we averaged across word embeddings of terms in that response.

\subsubsection{Document Embeddings}
The document embedding (also known as Doc2Vec) approach \cite{le_distributed_2014} to modelling text assigns n-dimensional vectors to variable-length textual content, such as sentences, paragraphs, and documents.
While it is closely related to the Word2Vec method of word embeddings, the document vectors are intended to represent the concept of a document as opposed to the context of a word in Word2Vec \cite{mikolov_distributed_2013}. Le and Mikilov \cite{le_distributed_2014} propose two Doc2Vec models, a distributed memory (Doc2Vec-DM) model and a distributed bag of words (Doc2Vec-DBOW) model. Doc2Vec-DM model is superior in terms of performance and usually achieves state-of-the-art results by itself. We used a Doc2Vec-DM model trained on content from Wikipedia to infer document vectors for candidate responses under this approach to modelling their language use.

\subsubsection{LIWC}
We also used the word categories from the Language Inquiry and Word Count (LIWC) lexicon \cite{pennebaker_development_2015}, the most popular closed-vocabulary approach used in linguistic analysis and modelling in the social science domains, especially for assessing personality-related constructs. LIWC 2015 version consists of 76 categories and the frequencies of occurrence of words in these categories in each candidate response normalized by the response length are used as features in modelling the language use.

\noindent Using the notation defined in (\ref{eq:tfidf}) and $ c $ denoting a category in LIWC lexicon, category frequency,
\begin{equation}
cf(c, r) =\dfrac{ \sum_{t \in c} n_{t, r}}{ \sum_{\acute{t} \in r} n_{\acute{t},r} }
\label{eq:liwc}
\end{equation}

\subsection{Text to Job-Hopping Motives Scale Model Building}
The above representations were used to build a regression model with the Random Forest \cite{breiman_random_2001} algorithm using the corresponding job-hopping motive scores as the target. We see as future work to compare the outcomes using different algorithms. We find it as sufficient to show the outcomes on a single algorithm in order to establish the correlation between language use and job-hopping motive as any improvement made over our findings using a different regression algorithm would only make the case for language-based inference of job-hopping motive stronger. 

We used 80\% of the data to train the model while the rest of the data was used to validate the accuracy of the trained model. 
We experimented with different minimum text response lengths, excluding records for candidates with responses shorter than the selected minimum word length.  The hypothesis behind this exercise was that responses that are too short might not have enough textual content to predict the candidates' job-hopping motive. We strived to find a balance between the minimum text response length and the data available for training the model to train the best predictive model. Table \ref{table:datasize} presents the number of records for different minimum response lengths.

\begin{table}
	\centering
	\caption{Data size for different minimum response length restrictions}
	\label{table:datasize}
	\begin{tabular}{| r | r |}
		\hline
		Min. response length (words) & Number of records (N) \\
		\hline
		50 & 45,899 \\ 
		100 & 32,472 \\ 
		150 & 23,675 \\
		200 & 18,210 \\
		\hline
	\end{tabular}	
\end{table}

\section{Results and Discussion}
\label{sec:results}

\begin{figure}[h]
	\centering
	\includegraphics[width=10cm]{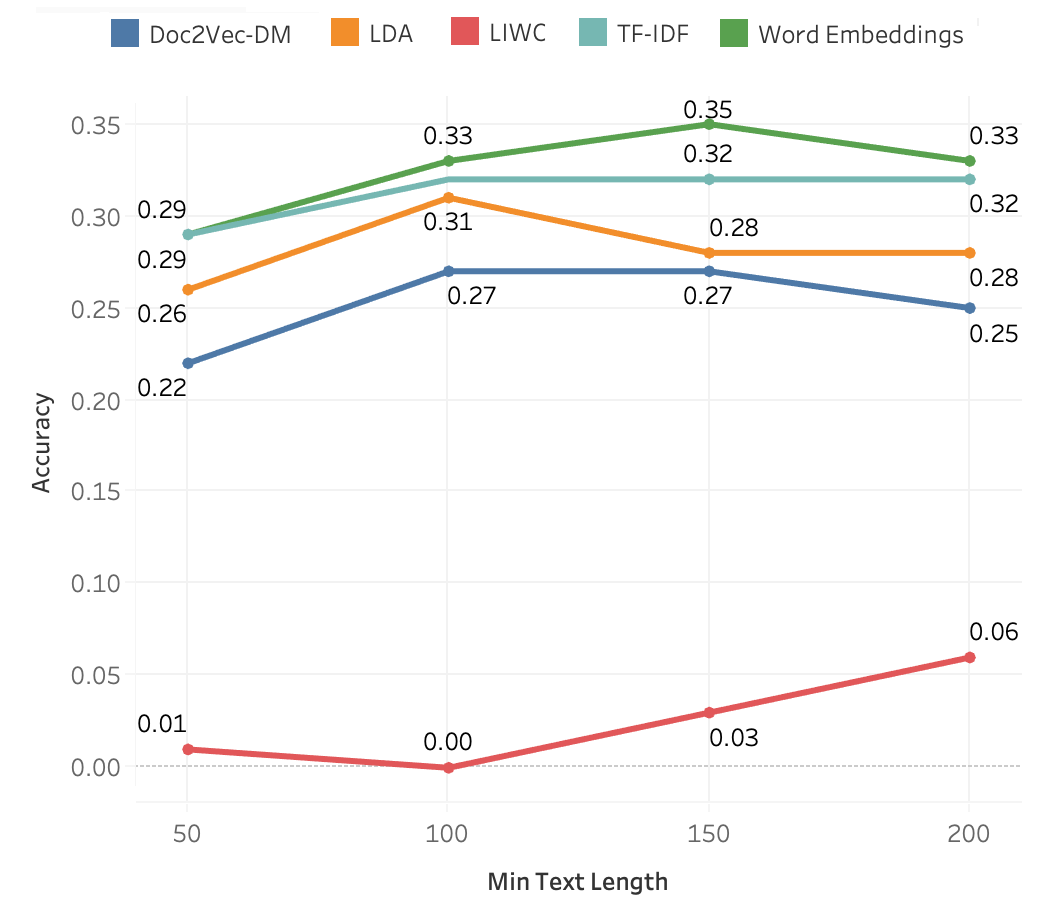}
	\caption{Model accuracy, measured in terms of the correlation coefficient between the job-hopping motive score (JHM Scale) and the score produced by the trained model, for different language modelling approaches and different minimum text lengths. All accuracies, except for LIWC with min. text lengths 50, 100 and 150, are significant at $ p = 0.001 $ level.}
	\label{fig:accuracy}
\end{figure}

We evaluated the trained ‘text to JHM Scale' regression models on the remaining 20\% of the data. The models are evaluated on the correlation coefficient between the actual JHM Scale score and the score predicted by the trained model. Figure \ref{fig:accuracy} shows the accuracies in terms of correlation across different minimum response lengths and language modelling approaches.

Language use representation using Glove word embeddings with minimum response length of 150 words achieved the highest correlation of r=0.35. It is important to note that six of the correlations fell above r=0.3, typically considered as a correlational upper-limit in personality research when predicting behaviour \cite{roberts_power_2007}. It is also important to note that apart from the correlations for LIWC with minimum text length of 50, 100 and 150 all other correlations had a $p < 0.001$. These results indicate that all open-vocabulary approaches across all minimum lengths recorded significant correlations, demonstrating that the language one uses in responding to typical interview questions are predictive of their job-hopping motive.

Overall, word embeddings based models recorded the highest corrections across all minimum lengths analysed. This is specifically important given that word embeddings based models are more generalizable to unseen words compared to models based on TF-IDF and LDA, which are limited to the vocabulary seen in the training corpus. The generalizability comes from the content used in training the word embedding model; In our case, the word embeddings used were trained on very generic content from web pages crawled from the Internet. 
Compared to the superior results achieved by word embedding based models, document embedding based models fell short of in terms of the accuracy. This, we believe, is due to the nature of the content used to train the document embedding model. We used content from Wikipedia to train our document embedding model and differences in actual content and writing style between Wikipedia and candidate responses may have contributed to the degraded performance of the document embedding based models. Further research is required to validate this and Doc2Vec models trained on other content such as tweets are options to consider.

Overall, minimum response length of 50 words achieved the weakest results confirming our hypothesis that responses that are too short might not have enough textual content to predict the candidates' job-hopping motive. However, none of the models showed an increase in accuracy but a decrease or maintaining the same accuracy (with the exception of LIWC based model),  when moving to a minimum length of 200 words. Given LIWC depends on counts of words related to pre-defined categories, we assume that more words in a response raise the possibility of finding more LIWC classified words in the response. However, the overall poor performance of LIWC highlights the limitations of closed-vocabulary approaches where a tediously developed lexicon is less effective in generalizing to unseen words. 

Following sections describe some of the further analysis we performed on the best ‘text to JHM Scale' model (min. response length=150, using word embeddings features) to get a deeper understanding of the model's behaviour. These analyses were carried out on the remaining 20\% of the data that were left out of training the regression model. The gender and role family composition of the this test data set can be found Tables \ref{table:likelihood} and \ref{table:gender}.

\subsection{Correlations with Personality and Language Characteristics}
\label{sec:correlations}

We evaluated the correlations between the output of the trained model (i.e. the inferred job-hopping motive) and candidates' personality measured in terms of the six-factor HEXACO trait model \cite{ashton_empirical_2007}. HEXACO, a six-factor model of personality developed by Ashton and Lee \cite{ashton_empirical_2007, lee_h_2013} is closely related to the Big Five model \cite{Goldberg1993TheSO} of personality but proposed as a better alternative, especially in explaining work-related behaviours. The six factors are \textit{Honesty-humility} (H), \textit{Emotionality} (E), \textit{eXtraversion} (X), \textit{Agreeableness} (A), \textit{Conscientiousness} (C) and \textit{Openness to experience} (O). We calculated each candidates HEXACO trait values using a language to HEXACO inference model described in \cite{Madhura_2020}.

\begin{table}
	\centering
	\caption{Correlations with the inferred job-hopping motive score. * $ p = 0.001 $}
	\label{table:correlations}
	\begin{tabular}{| l | r |}
		\hline
		Variable & Correlation coefficient \\
		\hline 
		HEXACO personality traits & \\ 
		\hspace*{0.5cm} Honesty-humility & -0.02~ \\  
		\hspace*{0.5cm} Emotionality & 0.06*  \\
		\hspace*{0.5cm} Extraversion & 0.01~ \\
		\hspace*{0.5cm} Agreeableness & -0.09* \\
		\hspace*{0.5cm} Conscientiousness & -0.00~ \\
		\hspace*{0.5cm} Openness to experience & 0.25* \\
		Response length in words & -0.15* \\
		Sentence count & -0.14* \\
		Formality score (F-score) & -0.22* \\
		Coleman Liau index & -0.16* \\
		Number of unique difficult words & -0.13* \\
		\hline
	\end{tabular}	
\end{table}

Table \ref{table:correlations} presents the correlations between the inferred job-hopping motive score and HEXACO personality traits, response length in words and number of sentences, Formality score (F-score) - a measure of formality and contextuality proposed by Heylighen and Dewaele \cite{heylighen_variation_2002}, Coleman Liau index - a commonly used measure of readability \cite{coleman_computer_1975} and count of ``difficult words'', identified by Dale and Chall \cite{dale_formula_1948} as words not included in a list of 3000 ``easy words'' commonly found in written English. F-score, Coleman Liau index and difficult words are measures of language proficiency and readability.

The negative correlations with response length and especially F-score indicate that candidates who are likely to hop jobs wrote less compared to others and used less sophisticated language. Moreover, the positive correlation of r=0.25 with the HEXACO \textit{Openness to experience} trait indicates candidates who are open to experiences are more likely to hop jobs. The positive correlation between \textit{Openness to experience} and turnover confirms what has been observed by Sarwar et al. \cite{sarwar_study_2013}, Zimmerman \cite{zimmerman_understanding_2008}, Timmerman \cite{timmerman_predicting_2006} and Anderson et al. \cite{woo_2011}. It is also interesting to note that the highest negative correlation (r=-0.09) with a HEXACO trait was recorded with \textit{Agreeableness}, a personality trait related to leniency in judging others, more willing to compromise and cooperate with others, and can easily control their temper \cite{ashton_empirical_2007}. The results indicate that personalities low in \textit{Agreeableness} are more likely to hop jobs.  

\subsection{Relationship with Candidate Demographics}

We inspected the relationship between the inferred job-hopping motive score and the job family the candidate applied to and their gender.

\begin{table}
	\centering
	\caption{Inferred job-hopping motive statistics for each job family}
	\label{table:likelihood}
	\begin{tabular}{| l | r | r |}
		\hline
		Job family & Count & Mean \\
		\hline  
		Cabin crew & 1,008 & 2.25 \\ 
		Call centre (outbound) & 207 & 2.38 \\ 
		Healthcare & 1,435 & 2.29 \\ 
		Retail & 852 & 2.37 \\ 
		Sales & 1,037 & 2.39 \\
		Other & 195 & 2.27 \\
		\hline
	\end{tabular}	
\end{table}

Table \ref{table:likelihood}  presents the statistics for each job family. The mean values for job families suggest that candidates for job families call centre (outbound), retail and sales have a higher tendency to hop jobs, which is in line with the general understanding of the job roles. Most of these are casual roles where candidate mobility is high, especially in outbound call centres. Sales roles are known to be stress causing related to target-driven nature of the operations. 
Moreover, an ANOVA analysis of job family data suggests that mean values demonstrate a statistically significant difference across job families.

Table \ref{table:gender} presents the statistics for gender. While the mean value for males is slightly higher than females', the effect size is 0.15 suggesting the difference is not significant. This is an important indication towards the trained model not showing bias towards any gender.

\begin{table}
	\centering
	\caption{Inferred job-hopping motive statistics for gender}
	\label{table:gender}
	\begin{tabular}{| l | r | r |}
		\hline
		Gender & Count & Mean \\
		\hline  
		Female & 1,339 & 2.31 \\ 
		Male & 1,348 & 2.33 \\
		Not specified & 2,047 & 2.32 \\
		\hline
	\end{tabular}	
\end{table}

\section{Conclusion and Future Work}
\label{sec:conclusion}
Frequent movement from job to job or ``job-hopping'' as its commonly known is found to be associated with one's personality. In this paper, we presented a novel approach to predicting job-hopping motive as measured by the Job-Hopping Motives (JHM) Scale using answers to typical interview questions related to past behaviour and situational judgement. Using data from over 45,000 individuals who answered open-ended interview questions and self-rated themselves using the JHM Scale, we built a regression model to infer the JHM Scale score. We compared the performance of four open-vocabulary text representation methods (namely terms, topics, word embeddings and document embeddings) and one closed-vocabulary method (LIWC). The Glove word embedding based model achieved the highest correlation of r=0.35 ($p < 0.001$) between interview response text and the JHM Scale score. All other open-vocabulary representations achieved correlations above 0.25 ($p < 0.001$), highlighting a statistically significant positive correlation between interview responses and job-hopping motive. We further demonstrated that one's job-hopping motive is positively correlated (r=0.25) with the trait \textit{Openness to experience} and negatively correlated with \textit{Agreeableness} (r=-0.09) as found in the six-factor HEXACO personality model. In other words, the more open someone is for new experiences and less lenient with views of others, the more likely he/she will show job-hopping motivation.   

We find the above outcome to be significant in at least two ways. It provides an alternative to resume based job histories as a source for inferring a job-applicants tendency to job hop. Resumes are known to induce bias in the hiring process and especially ineffective with newcomers to the job market with no significant prior job history. Secondly, the ability to infer job-hopping motive computationally from interview responses uplifts the utility of the interview as a multi-measure assessment that can be conducted digitally (e.g.text chat, video) at scale and cost-effectively giving every candidate an opportunity to express themselves. Interview as an assessment is preferred by applicants over traditional assessments such as personality tests.

Further work is required in assessing the predictive validity of the outcome, i.e. establishing the correlation between the inferred job-hopping motive scale score and actual job-hopping behaviour. This requires a longitudinal study or following the career journey of an applicant sample. In our current study, we used only the semantic level features (terms, topics etc). Exploring whether other types of features can further increase the accuracy, is another useful future extension. These can include the use of parts of speech (POS), use of emojis and multi-modal information such as audio and video signals captured while candidates answer the questions. Exploring the performance of other available regression algorithms, including neural network approaches, and using more advanced language representations such as BERT \cite{bert_2018}, may help increase the accuracy of the regression model further.

%\begin{acknowledgements}
%\end{acknowledgements}

\section*{Conflict of interest}
Both authors are employed at PredictiveHire, the creator of the FirstInterview product that was used to collect the data for the research.



\end{document}